%% file: kiev_paper.tex
\pdfoutput=1
\documentclass[runningheads]{llncs}
\usepackage{graphicx}
\usepackage{amsmath,amssymb} % define this before the line numbering.
\usepackage{times}
\usepackage{lineno}
\usepackage{color}
\usepackage[utf8]{inputenc}
\usepackage[T1]{fontenc}
\usepackage{subfigure}

\graphicspath{{bilder/}}
\newcommand{\DF}{{\ensuremath{\text{1} \hspace{-2.5pt} \text{I}}}}
\providecommand{\abs}[1]{\lvert#1\rvert}
\providecommand{\norm}[1]{\lVert#1\rVert}
\providecommand{\scalp}[2]{\left<#1,#2\right>}
\DeclareMathOperator*{\argmax}{arg\,max}
\begin{document}
%\renewcommand\thelinenumber{\color[rgb]{0.2,0.5,0.8}
%\normalfont\sffamily\scriptsize\arabic{linenumber}\color[rgb]{0,0,0}}
%\renewcommand\makeLineNumber {\hss\thelinenumber\ \hspace{6mm}
%\rlap{\hskip\textwidth\ \hspace{6.5mm}\thelinenumber}} 
%\linenumbers
\pagestyle{headings}
%==============================================================================
%\mainmatter
\title{Modelling Distributed Shape Priors by Gibbs Random Fields of Second
Order}
\titlerunning{Modelling Shape Priors by GRFs}

\author{Boris Flach\inst{1} \and Dmitrij Schlesinger\inst{2}}
\institute{Center for Machine Perception, Czech Technical University in Prague
\and Institute for Artificial Intelligence, Dresden University of Technology}

\maketitle

%%%%%%%%% ABSTRACT
\begin{abstract}
We analyse the potential of Gibbs Random Fields for shape prior modelling.
We show that the expressive power of second order GRFs is already sufficient to
express simple shapes and spatial relations between them simultaneously. This
allows to model and recognise complex shapes as spatial compositions of simpler
parts.
\end{abstract}

%%%%%%%%% BODY TEXT
%==============================================================================
\section{Introduction}
\subsubsection*{Motivation and goals}
Recognition of shape characteristics is one of the major
aspects of visual information processing. Together with colour, motion and
depth processing it forms the main pathways in the visual cortex.

Experiments in cognitive science show in a quite impressive way, that humans
recognise complex shapes by decomposition into simpler parts and interpreting
the former as coherent spatial compositions of these parts
\cite{Hoffmann:VI2000}. Corresponding guiding principles for the decomposition
where identified from these experiments as well as from research in computer
vision (see e.g.~\cite{Liu:CVPR2010}). The formulation of these principles
relies however on the assumption that the objects are already segmented and
thus concepts like convexity and curvature can be applied.

From the point of view of computer vision it is desirable to use shape
processing and modelling in the early stages of visual processing. This allows
to control e.g.~segmentation directly by prior assumptions or by feedback from
higher processing layers. This leads to the question whether composite shape
models can be represented and learned in a topologically fully distributed way.
The aim of the presented work is to study this question for probabilistic
graphical models. 
\subsubsection*{Related work}
All mathematically well principled shape models for early vision can be roughly
divided into the following two groups.

{\em Global models} treat shapes as a whole. Prominent representatives are
variational models and level set methods in particular. A shape is described up
to its pose by means of a level set function defined on the image domain.
Cremers et.al.~have shown in \cite{Cremers:IJCV06} how to extend these models
for scene segmentation. Recently we have shown how to use level set methods in
conjunction with MRFs \cite{Flach:SSPR08}. Global shape models are well suited
e.g.~for segmentation and tracking if the number of objects is known in advance
and a good initial pose estimation is provided.

{\em Semi global models} consider shape characteristics in local neighbourhoods
and go back to the ideas of G. Hinton on ``product of experts'' as well as of
Roth and Black on ``fields of experts'' (see \cite{Hinton:NC2002,Roth:IJCV09}
and citations therein). Mathematically these models are higher order GRFs of a
certain type -- additional auxiliary variables are used to express mixtures of
local shape characteristics in usually overlapping neighbourhoods.
Marginalisation over these auxiliary variables results in GRFs of higher order.
The work of Kohli, Torr et.al.~\cite{Ramalingam:CVPR08,Ladicky:ICCV2009}
demonstrates how to introduce such higher order Gibbs potentials directly and to
use them for segmentation in hierarchical Conditional Random Fields (CRF).
However, it is not clear how to learn the graphical structure for such models.
\subsubsection*{Contributions}
We will show that Gibbs Random Fields of second order have
already sufficient expressive power to model complex shapes as coherent spatial
compositions of simpler parts. Obviously, these models have to have a
significantly more complex graphical structure than just simple lattices.
Moreover, the graphical structure itself becomes a parameter which has to be
learnt together with the Gibbs potentials for each considered shape class.

From the application point of view these models have advantages especially in
the context of scenes with an unknown number of similar objects (i.e.~all
objects are instances of a single shape class). Moreover, such models can be
easily combined for scenes with instances of different shape classes.

The structure of the paper is as follows. In section~2 we introduce the GRF
model for composite shapes  and discuss the inference and learning tasks. The
latter means to learn the Gibbs potentials and the graphical structure itself. 
Section~3 gives experiments exploring the expressive power of the model -- first
we separately show its ability to express spatial relations between segments
and its ability to model simple shapes. Then we demonstrate its capability to
model composite shapes including structure learning. Finally, we show how
to combine such models for the discrimination of shape classes.
%==============================================================================
\section{The shape model}
\subsubsection*{Probability distribution}
\begin{figure}[t]
\begin{center}
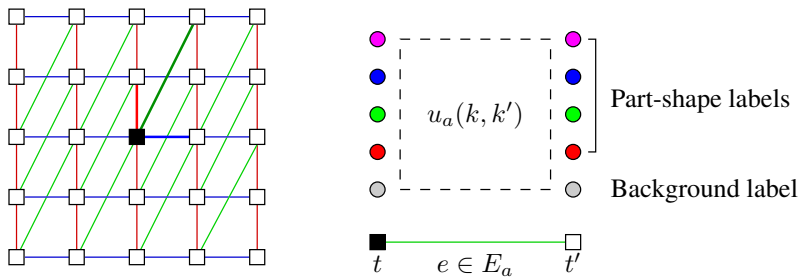
\end{center}
   \caption{Left: example of a translational invariant graphical structure.
Equivalence classes of edges $E_a$ are coloured by different colours. The set
$A$ is represented by bold edges outgoing from the central node. Right: Gibbs
potentials for an edge from $E_a$.}
\label{fig:graph}
\end{figure}

We begin with the description of the prior part of our shape model. Let
$D\subset\mathbb{Z}^2$ be a finite set of nodes $t\in D$, where each node
corresponds to an image pixel. Let $A\subset\mathbb Z^2$ be a set of vectors
used to define a neighbourhood structure on the set of nodes, i.e.~a graph: two
nodes $t$ and $t'$ are connected by an edge if $t'-t = a \in A$. To avoid double
edges we require $-A \cap A = 0$ (we use unary potentials as well). The
resulting graph is obviously translational invariant and the elements of $a\in
A$ define subsets $E_a\subset E$ of equivalent edges, where $e=(t,t') \in E_a$
if $t'-t=a$. A simple example is shown in Fig.~\ref{fig:graph}.

Given a class of composite shapes, we denote the set of its parts enlarged by an
extra element for the background by $K$.  A shape-part labelling $y\colon
D\rightarrow K$ is a mapping, that assigns either a shape-part label or the
background label $y_t\in K$ to each node $t\in D$. A function $u_a:K\times
K\rightarrow \mathbb{R}$ is defined for each difference vector $a\in A$. Its
values $u_a(k,k')$ are called Gibbs potentials. A corresponding probability
distribution is defined over the set of shape-part labellings as follows
\begin{equation}\label{eq:prior}
	p(y)=\frac{1}{Z(u)}\exp \sum_{a\in A} \sum_{tt'\in E_a}
	u_a\bigl(y_t,y_{t'}\bigr) ,
\end{equation}
where $Z$ denotes the partition sum (we omit the unary terms for better
readability). This p.d.~is homogeneously parametrised -- all edges in an
equivalence class $E_a$ have the same potentials.

\begin{remark}
Note that the parameters $u_a$ of this model are unique up to additive constants
for a given p.d.~under fairly general assumptions -- the only possible
equivalent transformations (aka re-parametrisations) consist in adding a
constant $\tilde{u}_a() = u_a() + \text{const}$. This will be shown in
appendix~\ref{app:unpot}. Therefore, we assume from here that the Gibbs
potentials for each $a\in A$ are normalised to sum to zero: $\sum_{k,k'}
u_a(k,k') = 0$. 
\end{remark}

\begin{remark}\label{rem:unpot}
It is important to notice that a homogeneously parametrised GRF on a finite
domain $D\subset\mathbb{Z}^2$ is not necessarily homogeneous. A p.d.~$p(y)$ for
labellings $y\colon D\rightarrow K$ is called homogeneous if its marginals for
congruent subsets coincide. This inhomogeneity, if present, usually reveals at
the domain boundary. It is easy to verify that the converse is true at least for
chains: a homogeneous Markov model on a finite chain admits a homogeneous
parametrisation.
\end{remark}

The appearance model is assumed to be a ``simple'' conditional independent
model. The probability to observe an image $x\colon D \rightarrow C$ ($C$ is
some colour space) given a shape-part labelling $y$ is
\begin{equation}\label{eq:appear}
 p(x \mid y ) = \prod_{t\in D} p\bigl(x_t\mid y_t\bigr) .
\end{equation}

In the light of the current popularity of CRFs it might well be asked, why we
decided to favour a GRF here. Both variants are identical with respect to
inference. Differences occur for learning.  We can imagine that shape-part
labellings can be used as latent variable layers for complex object segmentation
models. Recently, empirical risk minimisation learning has been proposed for
structured SVM models with latent variables \cite{Yu:ICML2009}. This shows that
learning of graphical models with latent variables is possible for both
variants -- GRFs and CRFs. However, since we want to study the expressive power
of the model in its pure form, we need a prior p.d.~and moreover, we want to be
able to learn such models fully unsupervised, which is possible for GRFs but not
for CRFs.
%==============================================================================
\subsubsection*{The inference task}
Informally, the inference task can be understood as follows. Given an
observation (i.e.~an image), it is necessary to assign values to all hidden
variables. We pose the segmentation task as a Bayesian decision task. Let $y'$
be the true (but unknown) segmentation and $C(y,y')$ be a loss function, that
assigns a penalty for each possible decision $y$. The task of Bayesian decision
is to minimise the expected loss
\begin{equation}
R(y ; x)=\sum_{y'} p(y'\mid x) C(y,y') \rightarrow \min_y .
\end{equation}
We use the number of misclassified pixels 
\begin{equation}
C(y,y')=\sum_t \DF\bigl\{y_t\neq y'_t\bigr\}
\end{equation}
as the loss function. It leads to the max-marginal decision
\begin{equation}\label{eq:recmarg}
y^\ast_t=\max_k p\bigl(y_t=k\bigm | x\bigr) \ \ \ \forall t\in D .
\end{equation}
Hence, it is necessary to calculate the marginal posterior probabilities for
each node $t\in D$ and label $k\in K$. Currently this task is infeasible for
GRFs. Several approximation techniques based e.g.~on belief propagation or
variational methods have been proposed for this task (see
e.g.~\cite{Wainwright:FTML08} for an overview). Unfortunately none of them
guarantees convergence to the exact values of the sought-after  marginal
probabilities. To our knowledge, the only scheme which does it is sampling,
which is however known to be slow \cite{Sokal:MCM1989}. 
%==============================================================================
\subsubsection*{Estimation of Gibbs potentials}
The learning task comprises to estimate the unknown model parameters given a
learning sample. We assume that the latter is a random realisation of i.i.d.
random variables, so that the Maximum Likelihood estimator is applicable.

The following situations are distinguished depending on the format of the
learning data. If the elements of the sample have the format $(x,y)$ then the
learning is called {\em supervised}. If, instead, they consist of images only
then the learning is called fully unsupervised. To cope with variants in-between
as well, i.e.~partial labellings $y_V$, we consider the elements of the training
sample to be events of the type $\mathcal{B}=(x,y_V) = \{(y,x) \mid y_{|V} =
y_V \}$.

We start with the learning of unknown potentials $u$. For simplicity we
consider the case when only one event $\mathcal{B}$ is given as the training
sample. According to the Maximum Likelihood principle, the task is
\begin{equation}\label{eq:ml0}
 p(\mathcal{B}; u)=\sum_{y\in\mathcal{B}} p(y) 
 p(x\lvert y) \rightarrow \max_u .
\end{equation}
Taking the logarithm and  substituting the model
\eqref{eq:prior}, \eqref{eq:appear} gives
\begin{equation}\label{eq:ml1}
 L(u) = \log \sum_{y\in\mathcal{B}} 
 \exp \Bigl[
 \sum_{a\in A} \sum_{tt'\in E_a}
 u_a\bigl(y_t,y_{t'}\bigr) \Bigr] p(x\lvert y) - 
 \log\bigl(Z(u)\bigr) \rightarrow \max_u .
\end{equation}
It is easy to prove, that the derivative with respect to the potentials is a
difference of expectations of some random variable $n_a(k,k';y)$ with
respect to the posterior and prior p.d.
\begin{equation}\label{eq:ml_grad_e}
 \partial L/\partial u_a(k,k') = 
 \mathbb{E}_{p(y\lvert\mathcal{B};u)}[n_a(k,k';y)]-\\
 \mathbb{E}_{p(y;u)}[n_a(k,k';y)] .
\end{equation}
The random variables $n_a(k,k';y)$ are defined by
\begin{equation}\label{eq:ml_grad_h}
 n_a(k,k';y)=\sum_{tt'\in E_a} \DF\bigl\{y_t{=}k,y_{t'}{=}k'\bigr\}
\end{equation}
and represent co-occurrences for label pairs $(k,k')$ along the edges in
$E_a$ for a labelling $y$. Combining these random variables into a random vector
$\Phi$, the gradient of the log-likelihood can be written as
\begin{equation}\label{eq:ml_grad_e_s}
 \nabla L(u) = \mathbb{E}_{p(y\lvert\mathcal{B};u)}[\Phi] - 
 \mathbb{E}_{p(y;u)}[\Phi] .
\end{equation}

The exact calculation of the expectations in \eqref{eq:ml_grad_e} is not
feasible. Therefore, we propose to use a stochastic gradient ascent to maximise
\eqref{eq:ml1}. The learning algorithm is an iteration of the following steps:
\begin{enumerate}
 \item Sample $\tilde{y}$ and $y$ according to the current
 a-posteriori probability $p(y|\mathcal{B};u)$ and a-priori probability
 $p(y;u)$ respectively.
 \item Compute $n_a(k,k';\tilde{y})$ and $n_a(k,k';y)$
 by \eqref{eq:ml_grad_h} for each $a\in A$, $k,k'\in K$.
 \item Replace the expectations in  \eqref{eq:ml_grad_e} by their
realisations and calculate new potentials $u$.
\end{enumerate}

For the sake of completeness we would like to mention that the learning of the
appearance models $p(c|k)$ can be done in a very similar manner. It is even
simpler from the computational point of view because the normalising constant
$Z$ does not depend on these probabilities. Therefore it is not necessary to
sample labellings according to the a-priori probability distribution $p(y)$.
Only a-posteriori sampled labellings are needed to perform the corresponding
stochastic gradient step.
%==============================================================================
\subsubsection*{Estimation of the interaction
structure}\label{subsec:strucestim}
A very important question not discussed so far is the optimal choice of the
neighbourhood structure $A$. Unfortunately, no well founded answer to this
question is known at present. One option is to use an abundant set of
interaction edges, e.g.~to assume that the set $A$ consists of all vectors
$A=\{a\in\mathbb{Z}^2 \mid \abs{a_1}, \abs{a_2}\leqslant d\}$ within a certain
range. Despite of the computational complexity this would lead to models with
high VC dimension and possibly -- as a result -- to weak discrimination. It is
therefore important to investigate the possibility to identify the neighbourhood
structure $A$ from a given training sample. A possible variant of a
corresponding formal task reads as follows. Given a training sample the task is
to find the best neighbourhood structure $A$ of given size $\abs{A}=m$ according
to the Maximum Likelihood principle $L(u_A,A) \rightarrow \max_{u_A,A}$. This
task is however not feasible - an exhaustive search over all possible sets $A$
would be computationally prohibitive, and, moreover, the likelihood can be
calculated only approximatively. Therefore we rely on a greedy approximation
which we will consider in two variants -- one of them successively includes new
elements into the neighbourhood structure starting from $A=\{0\}$  and the
other successively removes elements from this structure starting from
$A=\{a\in\mathbb{Z}^2 \mid \abs{a_1}, \abs{a_2}\leqslant d\}$.

For the first variant we use a greedy search for the interaction edges proposed
by Zalesny and Gimel'farb in the context of texture modelling
\cite{Zalesny:CVPR01,Gimelfarb:PAMI96}. Starting from the set $A=\{0\}$, i.e.~a
model with unary potentials, new edges are iteratively chosen and included into
$A$ as follows. First, the optimal set of potentials $u^*_A \in \mathcal{U}_A$
is determined for the current set $A$ as described in the previous subsection.
Here $\mathcal{U}_A$ denotes the subspace of potentials on the edges in $A$ (we
may assume that the Gibbs potentials are zero on all other edges). If a bigger
neighbourhood $A'$ is considered, then clearly, the gradient of the (log)
likelihood with respect to $u_{A'}$ in the point $u^*_A $ will be orthogonal to
the subspace $\mathcal{U}_A$. The proposal is to include the vector $a'\in
A'\setminus A$ with the largest gradient component
\begin{equation} \label{eq:struct_learn}
 a' = \argmax_{a\in A'\setminus A} \sum_{k,k'} \bigl[
 n_a(k,k';\mathcal{B},u) - n_a(k,k';u) \bigr]^2
\end{equation}
Optionally the Kullback-Leibler divergence can be used instead of the Euclidean
distance.

The second variant of structure estimation proceeds in opposite order. Starting
with the neighbourhood structure $A=\{a\in\mathbb{Z}^2 \mid \abs{a_1},
\abs{a_2}\leqslant d\}$, elements of $A$ are successively removed. The aim is to
remove in each step the element with the smallest impact on the maximal
likelihood
\begin{equation} \label{eq:changeLik}
 \max_{u_A} L(u_A) - \max_{u_{A\setminus a}} L(u_{A\setminus a})
 \rightarrow \min_{a\in A} .
\end{equation}
It is impossible to estimate this expression in the point $u_A^* = \argmax_{u_A}
L(u_A)$ using the gradient of the likelihood (like in the first variant) because
of $\nabla L(u_A^*) = 0$. It is nevertheless possible to estimate this
expression based on $u_A^*$. For the sake of simplicity we show this for the
situation of supervised learning. The likelihood maximisation with respect to
the Gibbs potentials reads
\begin{equation}\label{eq:maxLik}
 \max_{u_A} \Bigl\{ \scalp{\psi_A}{u_A} - 
 \log \sum_y \exp \scalp{\phi_A(y)}{u_A} \Bigr\}
\end{equation}
for this case. Here we have used the following notations. The set of all Gibbs
potentials $u_a(.,.)$, $a\in A$ is considered as a vector $u_A$. A realisation
of the random vector $\Phi_A$ (see \eqref{eq:ml_grad_e_s}) is denoted by
$\phi_A(y)$. Finally, $\psi_A$ denotes the corresponding vector of statistics
resulting from the training sample. Designating $\log Z(u_A)$ by $H(u_A)$, the
expression in \eqref{eq:maxLik} is nothing but the Fenchel conjugate
$H^*(\psi_A)$. It is known that for exponential families the latter can be
written as
\begin{equation}
 H^*(\psi_A) = \inf \Bigl\{
 \sum_y p(y)\log p(y) \Bigm | \mathbb{E}_p[\Phi_A] = \psi_A ,
 \hspace{.3em}
 p \in \mathcal{P} \Bigr\}
\end{equation}
(see~e.g.~\cite{Wainwright:FTML08,Borwein:CANO2000}), where we denoted the
expectation w.r.t.~a probability distribution $p$ by $\mathbb{E}_p$ and the set
of all probability distributions on labellings $y$ by $\mathcal{P}$. This means
to find the p.d.~with maximal entropy among {\em all} distributions having
expectation $\psi_A$ of the random vector $\Phi_A$.

Removing an element $a$ from the neighbourhood structure $A$ can be equivalently
expressed by the linear constraints $u_a \equiv 0$. Considering the task
\eqref{eq:maxLik} with these additional constraints, it can be shown by
the use of Fenchel duality (see e.g.~\cite{Borwein:CANO2000}) that the
corresponding conjugate function $\widetilde{H}^*(\psi_A)$ can be written as
\begin{equation}
 \widetilde{H}^*(\psi_A) = \inf_{z_a} \inf_p \Bigl\{
 \sum_y p(y)\log p(y) \Bigm | \mathbb{E}_p[\Phi_A] = \psi_A + z_a,
 \hspace{.3em}
 p \in \mathcal{P} \Bigr\},
\end{equation}
where $z_a$ is an arbitrary vector of the subspace $\mathcal{U}_a$. Therefore,
the difference in \eqref{eq:changeLik} is equal to $H^*(\psi_A) -
\widetilde{H}^*(\psi_A) = H^*(\psi_A) - H^*(\psi_A + z_a^*)$ and can be
estimated by the gradient of $H^*$ in $\psi_A$. The latter gradient is nothing
but the vector of Gibbs potentials $u_A^*$.

\begin{remark}
The convex, lower semi-continuous function $H^*(\psi_A)$ is not differentiable
in general. Therefore its sub-differential may consist of more than one
subgradient $u_A$. This corresponds to the non-uniqueness of the Gibbs
potentials. We have however shown that the Gibbs potentials are unique up to
additive constants for the model class considered in this paper (see
Remark~\ref{rem:unpot} and Appendix~\ref{app:unpot}). 
\end{remark}

Summarising, the difference in \eqref{eq:changeLik} can be estimated by
$\norm{u_a}$, what leads to the following greedy removal strategy for 
elements of the neighbourhood structure $A$. Given a current neighbourhood
structure $A$, estimate the optimal Gibbs potentials $u_A^*$ and remove the
the element $a\in A$ with the smallest value of $\norm{u_a}$.
%==============================================================================
\section{Experiments}
%==============================================================================
\subsubsection*{Modelling spatial relations between segments}

\begin{figure}[t]
\begin{center}
\includegraphics[width=0.175\textwidth]{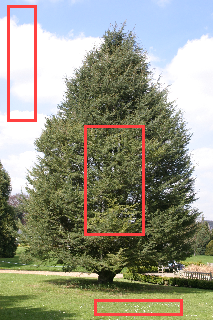}
\includegraphics[width=0.175\textwidth]{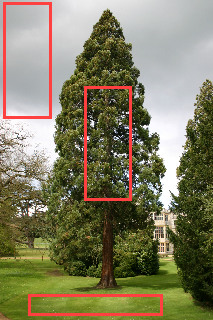}
\includegraphics[width=0.175\textwidth]{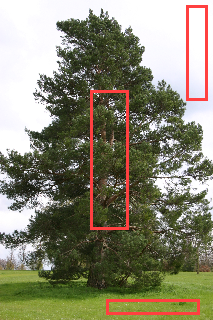}

\vspace{0.002\textheight}
\includegraphics[width=0.175\textwidth]{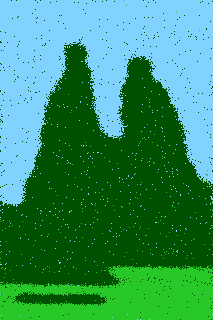}
\includegraphics[width=0.175\textwidth]{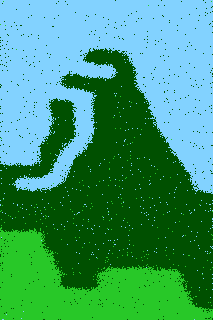}
\includegraphics[width=0.175\textwidth]{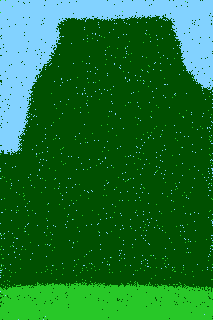}

\vspace{0.002\textheight}
\includegraphics[width=0.175\textwidth]{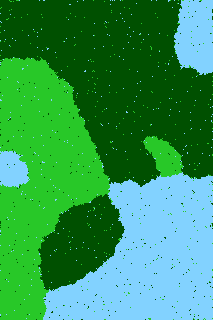}
\includegraphics[width=0.175\textwidth]{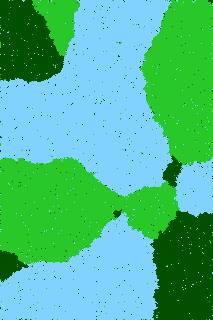}
\includegraphics[width=0.175\textwidth]{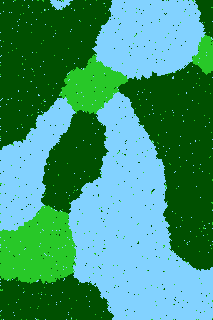}
\end{center}
\caption{\label{fig:trees_1}Modelling spatial relations between segments. The
first row shows input images and regions with fixed segmentation. The middle and
bottom row show labellings generated by the learned a-priori models (segment
labels are coded by colour): the images in the middle row were generated by the
model with full neighbourhood, whereas the images in the bottom row were
generated by the baseline model.}
\end{figure}

The first experiment investigates the  ability of the model
\eqref{eq:prior}, \eqref{eq:appear} to reflect  spatial relations between
segments, i.e.~scene parts, which are too large to capture their shape by a
neighbourhood structure of reasonable size. We used the three images shown in
the first row of Fig.~\ref{fig:trees_1} as training examples. Each scene should
be segmented into three segments: $K=\{sky, trees, grass\}$. The appearance
models $p(c\lvert k)$ for the segments were assumed as mixtures of multivariate
Gaussians (four per segment). A model with ''full`` neighbourhood structure  --
all vectors $\{a\in\mathbb Z^2 \mid |a_1|, |a_2| \leq d\}$ with $d=20$ was used
in this experiment. A ``simple'' but anisotropic Potts model on the
8-neighbourhood was chosen as a baseline for comparison.

Semi-supervised learning was applied by fixing the segment labels in the
rectangular areas shown by red rectangles during learning. Both the a-priori
models (the potentials and the direction specific Potts parameters for the
baseline model) and the appearance models (mixture weights, mean values and
covariance matrices) were learned. 

The difference of the models can be clearly seen by observing labellings
generated a-priori by the learned models, i.e.~without input images. Some of
them are shown in the second and third row for the model with complex
neighbourhood structure and the baseline model respectively. It can be seen,
that the spatial relations between segments (like e.g.~``above'', ``below''
etc.) were correctly captured by the complex model, whereas it is clearly not
the case for the Potts model.

\begin{figure}[t]
\begin{center}
\includegraphics[width=0.2\textwidth]{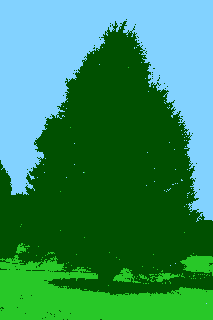}
\includegraphics[width=0.2\textwidth]{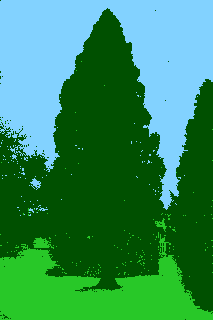}
\includegraphics[width=0.2\textwidth]{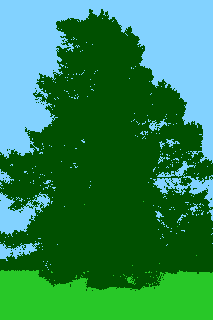}

\vspace{0.002\textheight}
\includegraphics[width=0.2\textwidth]{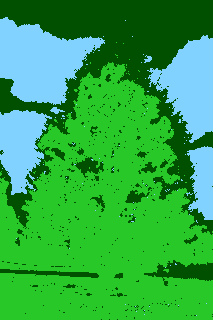}
\includegraphics[width=0.2\textwidth]{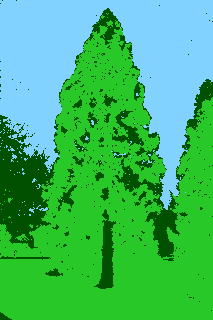}
\includegraphics[width=0.2\textwidth]{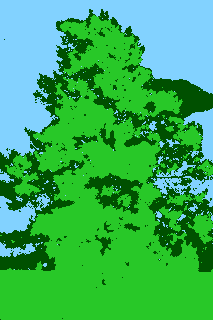}
\end{center}
\caption{\label{fig:trees_2}Segmentation results obtained after fully
unsupervised learning of the appearance part of the model. Upper row -- model
with full neighbourhood, bottom row -- baseline model.}
\end{figure}

The consequences can be clearly seen from the following experiment. We fixed the
prior models obtained in the previous experiment (semi-supervised learning) for
both variants (the complex prior and the Potts prior) and learned the parameters
of the Gaussian mixtures completely unsupervised. Fig.~\ref{fig:trees_2} shows
labellings (i.e.~segmentations) sampled at the end of the learning process by
the corresponding a-posteriori probability distributions (obtained with the
learned appearances) for the complex a-priori model and the Potts a-priori model
in the first and the second row respectively. The advantages of the complex
model are clearly seen. These results can be explained as follows. There are
twelve Gaussians in total to interpret the given images. For the learning
process it is ``hard to decide'' which of the Gaussians belongs to which
segment. Using the compactness assumption only, is obviously not enough to
separate segments from each other. If the complex model is used instead, the
learning process starts to generate labellings according to the a-priori
probability distribution, i.e.~labellings which reflect the correct spatial
relations between the segments. This forces the unsupervised learning of the
appearance models into the right direction.
%==============================================================================
\subsubsection*{Modelling simple shapes}

\begin{figure}[t]
\begin{center}
\includegraphics[width=0.25\textwidth]{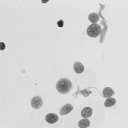}
\includegraphics[width=0.25\textwidth]{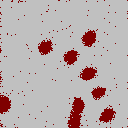}

\vspace{0.002\textheight}
\includegraphics[width=0.25\textwidth]{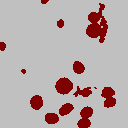}
\includegraphics[width=0.25\textwidth]{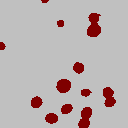}
\end{center}
\caption{\label{fig:cells}Modelling and segmentation of simple shapes. Upper
left -- input image, upper right -- a labelling generated a-priori by the
learned complex model. Final segmentations are shown in the bottom row: left --
baseline model, right -- complex model.}
\end{figure}

This group of experiments demonstrates the ability of the model to represent
simple shapes as well as to perform shape driven segmentation. This experiment
is prototypical e.g.~for a class of image recognition tasks in biomedical
research. Fig.~\ref{fig:cells} (upper left) shows a microscope image of liver
cells with stained DNA. Thus, only the cell nuclei are visible.  The task is to
segment the image into two segments -- ''cells`` (which have nearly circular
shape) and ''background`` (the rest including artefacts). Hence, two labels are
used. The ''full`` neighbourhood structure with $d=12$ was used (it
approximately corresponds to the mean cell diameter). Again, we used a baseline
model for comparison -- a GRF with 4-neighbourhood and free potentials. The
appearances for grey-values were assumed to be Gaussian mixtures (two per
segment) in both models.

First, semi-supervised learning was performed (like in the previous experiment
with trees) in order to learn the prior distributions for labellings  as well as
the appearances for both, the complex and the baseline model. A labelling
generated a-priori by the learned complex model is shown in Fig.~\ref{fig:cells}
(upper right). The final segmentations according to the max-marginal decision
(see equation \eqref{eq:recmarg}) are shown in the bottom row of the same
figure. The differences are clearly seen. The shape prior captured in the
complex model led to the correct segmentation -- the artefacts were segmented as
background, whereas the baseline model produces a wrong segmentation because
neither the appearance nor a simple ''compactness`` assumption nor even their
combination allow to differentiate between cells and artefacts. 
%==============================================================================
\subsubsection*{Structure estimation for simple shapes}
\begin{figure}[t]
\begin{center}
\includegraphics[width=0.3\textwidth]{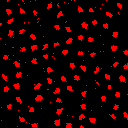} 
\includegraphics[width=0.41\textwidth]{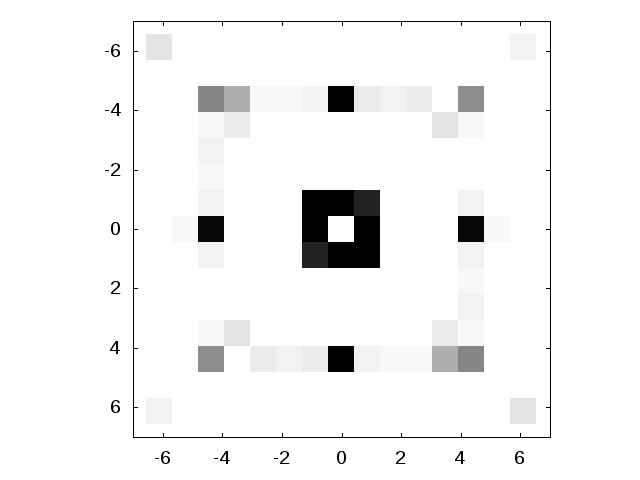}
\end{center}
\caption{\label{fig:blobs}Shape estimation for a simple shape model. Left --
labelling generated by the known model, right -- histogram of the estimated
structures.}
\end{figure}

In order to investigate the structure identifiability of shape models we have
used an artificial model which generates simple ''blobs``. The neighbourhood
structure consists of 8 elements. The group of the first four elements with
coordinates $(0,1)$, $(0,-1)$, $(1,1)$ and $(-1,1)$ describes a standard
8-neighbourhood. The remaining four vectors are scaled versions of the first
(scale factor 5). The Gibbs potentials on the short vectors are supermodular and
express the correlation of the labels on the edges of this type
\begin{equation}
 u(k,k') = \begin{cases}
              \alpha & \text{if $k=k'$,} \\
              -\alpha & \text{else.}
             \end{cases} .
\end{equation}
The Gibbs potentials on the long edges consist of an submodular and a modular
part $u(k,k') = u_1(k,k') + u_2(k,k')$, where the submodular part $u_1$ is
just the negative version of the potentials on the short edges and expresses an
anti-correlation of the labels on these edges. The modular part
\begin{equation}
 u_2(k,k') = \begin{cases}
              \beta & \text{if $k=k'=0$,} \\
              -\beta & \text{if $k=k'=1$,} \\
              0 & \text{else}
             \end{cases} .
\end{equation}
is used to influence the density of the blobs. A labelling (fragment) sampled by
this model ($\alpha=0.35$, $\beta=0.5$) is shown in Fig.~\ref{fig:blobs}. Both
heuristic approaches for structure estimation discussed in the previous section
where applied for the supervised version, i.e.~using a labelling generated by
the known model as a learning sample.

The first approach -- iterative growth of the structure -- was run 40 times. The
estimated structures resulting from these runs are shown in Fig.~\ref{fig:blobs}
as a grey-coded histogram. As a stochastic gradient ascend is used for the
learning of the potentials, each run may result in a different structure. The
histogram shows however, that the structure estimation is essentially correct. 
All trials of the second approach -- iterative shrinking of the neighbourhood
structure -- resulted much to our surprise in one and the same estimated
structure -- the correct one.

We conclude from these experiments that the neighbourhood structure of a shape
model is identifiable (at least in principle) from labellings generated by the
model.

%==============================================================================
\subsubsection*{Modelling composite shapes}
\begin{figure}[t]
\begin{center}
\includegraphics[width=0.25\textwidth]{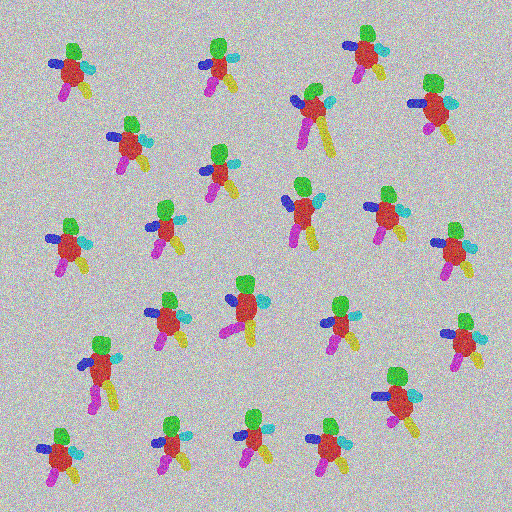}
\includegraphics[width=0.25\textwidth]{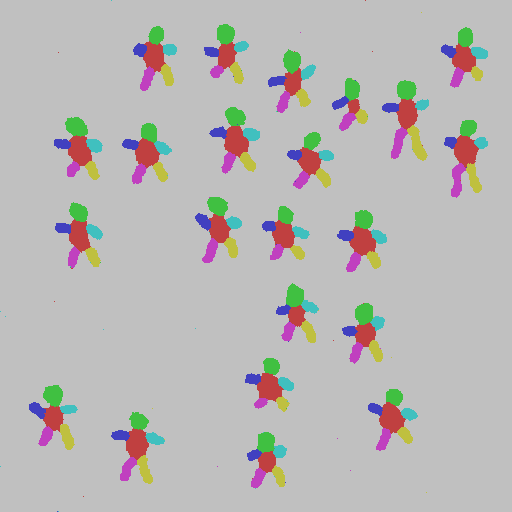}
\includegraphics[width=0.25\textwidth]{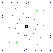}

\vspace{0.002\textheight}
\includegraphics[width=0.25\textwidth]{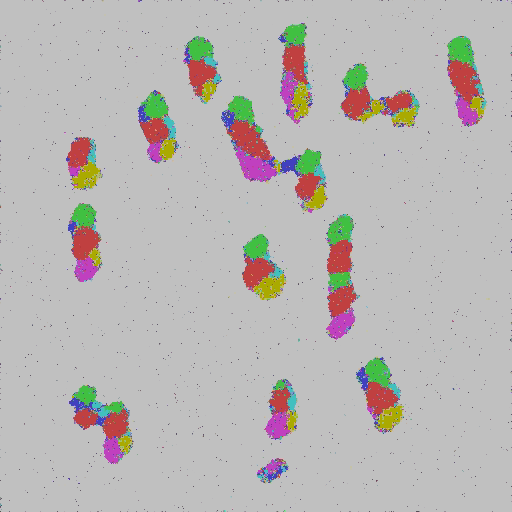}
\includegraphics[width=0.25\textwidth]{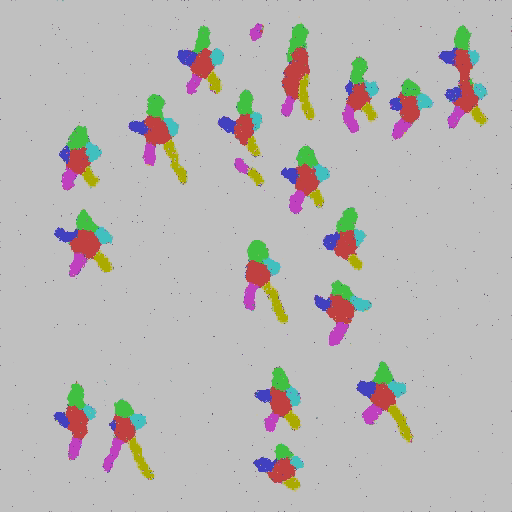}
\includegraphics[width=0.25\textwidth]{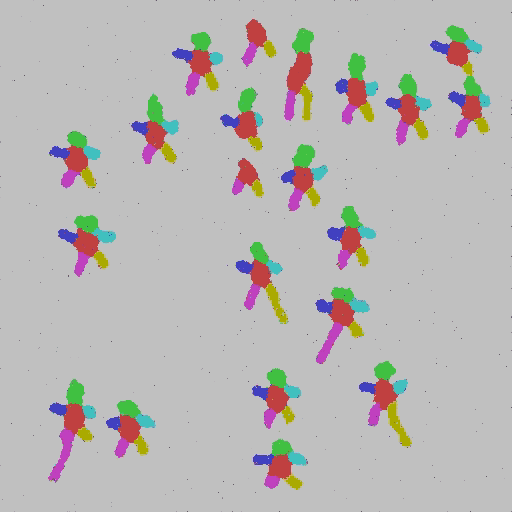}
\end{center}
\caption{\label{fig:man}Composite shape modelling. Upper row from left: input
image, labelling generated a-priory by the learned model, estimated interaction
structure. Bottom row: labellings generated by models during learning.}
\end{figure}

The previous experiments have shown that second order GRFs can model both,
spatial relations between segments and simple shapes. Now we are going to
demonstrate the capability of the model to capture both properties {\em
simultaneously}. This opens the possibility to represent complex shapes as
spatial compositions of simpler parts. To demonstrate this, we use an artificial
example shown in Fig.~\ref{fig:man} (upper left). It was produced manually and
corrupted by Gaussian noise. Accordingly, the model was defined as follows. The
label set $K$ consists of seven labels, each one corresponding to a part of the
modelled shape (as well as one for the background). The appearance models
$p(c\lvert k$) for the labels are Gaussians with known parameters. In this
experiment we applied the growth variant for the estimation of the interaction
structure as described in section \ref{subsec:strucestim}. 

Fig.~\ref{fig:man} (upper row, center) shows a labelling generated by the
learned prior model. It is clearly seen that  both, spatial relations between
object parts and part shapes are captured correctly. 

The bottom row of Fig.~\ref{fig:man} displays labellings generated during the
process of structure learning at time moments, when the interaction
structure learned so far was not yet capable to capture all needed properties.
As it can be seen, the model was able to learn spatial relations between
the segments more or less correctly even for a small numbers of edges ($5$
edges -- bottom left). More relations are learned as the number of edges grows
(bottom middle and right). Finally, $20$ difference vectors were necessary to
capture all relations (out of $1200$ possible for the maximal range of $d=24$).

Fig.~\ref{fig:man} (upper right) shows the estimated neighbourhood structure.
The endpoints of all edges from central pixel are marked by colours (the image
is magnified for better visibility). A certain structure can be seen in this
image. The $8$-neighbourhood edges (black) reflect compactness and adjacency
relations of the object parts. The learned potentials on these edges represent
strong label co-occurrences. Most of the other vectors are responsible for the
shapes of the parts. The potentials on the red edges express characteristic
breadths, and the potentials on the green edges -- characteristic lengths of the
parts. The potentials on these edges mainly represent anti-correlations, forcing
label values to change along  certain directions. The blue pixels in the figure
reflect relative positions of object parts.

%==============================================================================
\subsubsection*{Composite shape recognition}

\begin{figure}[t]
\begin{center}
\includegraphics[width=0.35\linewidth]{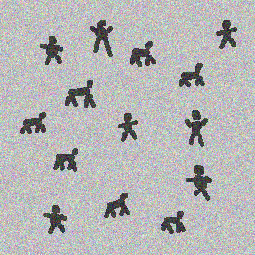} 
\includegraphics[width=0.35\linewidth]{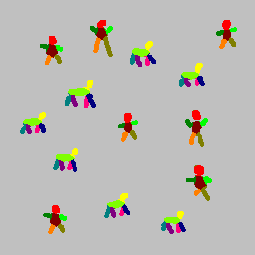}
\end{center}
\caption{\label{fig:catman}Shape segmentation and classification. Left -- input
image, right -- segmentation (part-labels are encoded by colours).}
\end{figure}

\begin{figure}[t]
\begin{center}
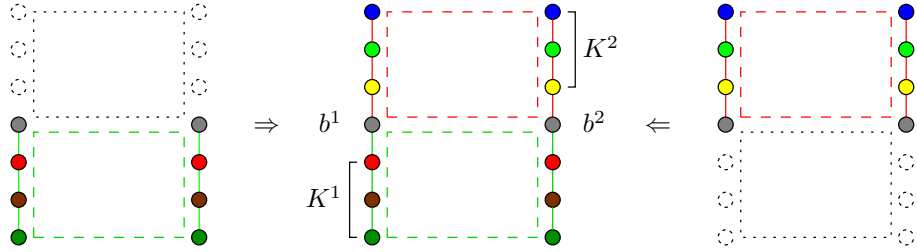
\end{center}
\caption{Estimation of the aposteriori statistics for the joint model. Left and
right: extended statistics for shape models. Middle: the joint model --
statistics marked green and red are inherited from the components. Others are
set to a small constant.}
\label{fig:compos}
\end{figure}

The final experiment demonstrates possibilities to combine composite shape
models. The aim is to obtain a joint model which can be used for detection,
segmentation and classification of objects in scenes populated by instances of
different shape classes like e.g.~the example in Fig.~\ref{fig:catman}. We
conclude from the previous experiments, that the appearance model can be
re-learned in a fully unsupervised way if the prior shape model is
discriminative. Hence, the most important question is, how to combine the prior
models. We propose a method for this that is based on the following observation.
It is not necessary to have an example image (or an example segmentation) in
order to learn the model if the aposteriori statistics
\begin{equation}\label{eq:statistics}
\bar\Phi_a(k,k')=\mathbb{E}_{p(y\lvert\mathcal{B},u)}\bigl[\Phi_a(k,k')\bigr]
\end{equation}
for all difference vectors $a\in A$ and label pairs $(k,k')$ are known -- the
gradient of the likelihood (equation \eqref{eq:ml_grad_e}) reads then
\begin{equation}\label{eq:grad1}
 \partial L/\partial u_a(k,k') = \bar\Phi_a(k,k') -
 \mathbb{E}_{p(y;u)}[n_a(k,k';y)] .
\end{equation}
Let us consider this in a bit more detail for a simple example -- just two
shapes like in Fig.~\ref{fig:catman}. Let us assume that the both models are
learned, i.e.~both the potentials and statistics are known for both models and
for all difference vectors $a$. Obviously, it is not easy to combine the
potentials of both shape models in order to obtain new ones for a model that
generates such collages. It is however very easy to estimate the needed
aposteriori statistics for the joint model given the aposteriori statistics for
both shape models. Summarizing, the scheme to obtain the parameters of the joint
model consists of two stages:
\begin{enumerate}
\item compute the aposteriori statistics for the joint model and
\item learn the model according to \eqref{eq:grad1} so that it reproduces
this statistics.
\end{enumerate}
As the second stage is standard, we consider the first one in more detail. Let
us denote the label sets corresponding to the shape parts by $K^1$ and $K^2$ for
the first and for the second shape type respectively. Let $b^1$ and $b^2$ be the
background labels in the corresponding shape models and $b$ be the background
label in the joint one. Consequently, the label set of the latter is $K^1\cup
K^2 \cup b$ (see the middle part of Fig.~\ref{fig:compos}). 

First of all we enlarge the label sets of each shape model by labels that are
not present in this model but present in the joint one. Thereby the statistics
for the new introduced labels (for all difference vectors $a$) are set to zero
(see Fig.~\ref{fig:compos}, left and right). Informally said, these extended
aposteriori statistics correspond to the situations that the joint model is
learned on examples, in which only labels of one particular shape are present.
The aposteriori statistics for the joint model is then obtained as a weighted
mixture of the two extended ones and an additional uniformly distributed
component. The latter is added in order to avoid zero probabilities (which
would lead to obvious technical problems for the Gibbs Sampler). Summarising,
the aposteriori statistics of label pairs for a difference vector $a$ of the
joint model is:
\begin{eqnarray}
\bar\Phi_a(k,k')\sim \left\{
\begin{array}{lll}
w_1\cdot \bar\Phi_a^1(k,k')+w_0 & \text{\ \ \ if} & k\in K^1 \text{\ and\ }
k'\in K^1
,\\
& & k\in K^1 \text{\ and\ } k'=b , \\
& & k=b \text{\ and\ } k'\in K^1 , \\
w_2\cdot \bar\Phi_a^2(k,k')+w_0 & \text{\ \ \ if} & k\in K^2 \text{\ and\ }
k'\in K^2
,\\
& & k\in K^2 \text{\ and\ } k'=b , \\
& & k=b \text{\ and\ } k'\in K^2 , \\
w_1\cdot \bar\Phi_a^1(b^1,b^1)+\\
\text{\ \ \ } + w_2\cdot \bar\Phi_a^2(b^2,b^2)+w_0 &
\text{\ \ \ if} & k=b \text{\ and\ } k'=b \\
w_0 & & \text{\ \ \ otherwise.}
\end{array}
\right. 
\end{eqnarray}
with some weights $w_0\ll w_1\approx w_2$, where the indices $1$ and $2$
correspond to the particular shape model. Given these statistics the joint model
is learned according to \eqref{eq:grad1}.

For the experiment in Fig.~\ref{fig:catman} two composite shape models were
learned separately. The test image in Fig.~\ref{fig:catman} (left) is a collage
of both shape types. Note that the appearance of all shape parts is identical,
so they are not distinguishable without the prior shape model.
Fig.~\ref{fig:catman} (right) shows the final segmentation. It is seen that all
objects were correctly segmented and recognised -- although both composite shape
classes share some similarly shaped parts -- they were not confused.
%==============================================================================
\section{Conclusions}
The notation of shape is often understood as an object property of global
nature. We followed a different direction by modelling shapes in a distributed
way. We have demonstrated that the expressive power of second order GRFs allows
to model spatial relations of segments, simple shapes and moreover, both aspects
{\em simultaneously} i.e.~composite shapes which are understood as coherent
spatial compositions of simpler shape parts.

We have shown that complex shapes can be recognized even in the situation, when
their parts are not distinguishable by appearance. However, in our learning
experiments we used training images, where they are distinguishable. Thus, an
important question is, whether it is possible to perform unsupervised
decomposition of complex shapes into simpler parts during the learning phase,
i.e.~to learn shape models from images, where the desired spatial relations
between shape parts are not explicitly present. Another important issue is the
learning of the interaction structure. It would be very useful to have a 
well grounded approach for this.

\subsection*{Acknowledgments}
We would like to thank Georgy Gimel'farb (University of Auckland) for the
fruitful and instructive discussions which have been particularly valuable with
regard to structure learning.

One of us (B.F.) was supported by the Czech Ministry of Education project
1M0567. D.S.~was supported by the Deutsche Forschungsgemeinschaft, Grant
FL307/2-1. Both authors were partially supported by Grant NZL 08/006 of the
Federal Ministry of Education and Research of Germany and the Royal Society of
New Zealand.

{\small
\bibliographystyle{splncs03}
\bibliography{kiev_paper}
}
%==============================================================================
\appendix
\section{Equivalent transforms for homogeneously parametrised
GRFs}\label{app:unpot}
As we have already seen, the probability distribution~\eqref{eq:prior} for shape
part labellings $y$ can be equivalently written as
\begin{equation}
 p(y) \sim \exp\Bigl[
 \sum_k u_0(k) n_0(k; y) + 
 \sum_{a\in A'} \sum_{kk'} u_a(k,k') n_a(k,k; y)
 \Bigr ],
\end{equation}
where $A' = A \setminus \{0\}$.
We call two parametrisations $u$, $\tilde{u}$ equivalent, if the corresponding
probability distributions are identical. It follows that the difference $v = u
- \tilde{u}$ of equivalent potentials fulfils
\begin{equation}
 V(y) = 
 \sum_k v_0(k) n_0(k;y) + 
 \sum_{a\in A'} \sum_{kk'} v_a(k,k') n_a(k,k'; y) = \text{const.}
\end{equation}
We will conclude that all functions $v_a$ are constant under fairly general
conditions. We perform the proof in two steps. First we show that the
pairwise functions $v_a$, $a\not = 0$ are modular and can be written as a sum of
unary functions. In a second step we will conclude the claimed statement under
fairly general conditions for the graph $(D,E)$.

Let us consider an arbitrary non-zero vector $a \in A$ of the neighbourhood
structure and an arbitrary edge $(tt')\in E_a$. Let $k_1,k_2$ be two arbitrary
labels in the node $t$ and $k'_1,k'_2$ be two arbitrary labels in the node
$t'$. Let $y_{11},y_{12},y_{21},y_{22}$ be four labellings with respective
values $(k_1,k'_1),(k_1,k'_2),(k_2,k'_1),(k_2,k'_2)$ on the nodes $t,t'$
such that they coincide on all other vertices. We consider the equation
\begin{equation}
 V(y_{11}) + V(y_{22}) - V(y_{12}) - V(y_{21}) = 0 .
\end{equation}
It is easy to see that this equation reduces to
\begin{equation}
 v_a(k_1,k'_1) + v_a(k_2,k'_2) - v_a(k_1,k'_2) - v_a(k_2,k'_1) = 0. 
\end{equation}
This holds for arbitrary four-tuples of labels and it follows that the
function $v_a$ is modular and can be written as a sum of two unary functions
\begin{equation}
 v_a(k,k') = \tilde{v}_a(k) + \tilde{v}_{-a}(k').
\end{equation}
These arguments can be applied for every element $a\in A'$. Consequently, $V(y)$
can be written as
\begin{equation}
  V(y) = 
 \sum_k v_0(k) n_0(k;y) + 
 \sum_{a\in A'}\sum_k \bigl[
 v_a(k) n_a(k;y) + v_{-a}(k) n_{-a}(k;y) 
 \bigr] ,
\end{equation}
where we have omitted the tildes. Note that $n_a(k;y) = \sum_k n_a(k,k';y)$
denotes the number of vertices with an outgoing edge of type $a$ for which the
labelling $y$ has the value $k$. Therefore in general $n_0(k;y) \not =
n_a(k;y)$.

Let us consider an arbitrary vertex $t$ and two labellings $y,\tilde{y}$ which
coincide on all vertices but $t$. It follows from $V(y) - V(\tilde{y}) = 0$
that
\begin{equation}\label{eq:sumun}
 v_0(k) + 
 \sum_{\substack{a\in A'\\ t+a\in D}} v_a(k) + 
 \sum_{\substack{a\in A'\\ t-a\in D}} v_{-a}(k) = \text{const.}
\end{equation}
We assign a vector $z(t)$ with dimension $2\abs{A}-1$ to every vertex
$t\in D$ with components
\begin{equation}
 z_0(t) = 1 ,\hspace{.3em}
 z_a(t) = \begin{cases}
           1 & \text{if $t+a\in D$,}\\
           0 & \text{else}
          \end{cases} \hspace{.3em}
          \text{and }
 z_{-a}(t) = \begin{cases}
           1 & \text{if $t-a\in D$,}\\
           0 & \text{else.}
          \end{cases}
\end{equation}
If the domain $D$ contains a subset of nodes $t$ such that their vectors
$z(t)$ span the whole vector space of dimension $2\abs{A}-1$, then, clearly,
considering equation \eqref{eq:sumun} for each of them, we obtain
\begin{align}
 v_0(k) = & \text{const.} \\
 v_a(k) = & \text{const.} \\
 v_{-a}(k) = & \text{const.} 
\end{align}
for all $a\in A$. $\qed$

\end{document}

%% file: bilder/graph.pdf_t
\begin{picture}(0,0)%
\includegraphics{graph.pdf}%
\end{picture}%
\setlength{\unitlength}{4144sp}%
\begingroup\makeatletter\ifx\SetFigFont\undefined%
\gdef\SetFigFont#1#2#3#4#5{%
  \reset@font\fontsize{#1}{#2pt}%
  \fontfamily{#3}\fontseries{#4}\fontshape{#5}%
  \selectfont}%
\fi\endgroup%
\begin{picture}(4719,1651)(439,-1250)
\put(3241,-286){\makebox(0,0)[b]{\smash{{\SetFigFont{10}{12.0}{\familydefault}{\mddefault}{\updefault}{\color[rgb]{0,0,0}$u_a(k,k')$}%
}}}}
\put(2656,-1186){\makebox(0,0)[b]{\smash{{\SetFigFont{10}{12.0}{\familydefault}{\mddefault}{\updefault}{\color[rgb]{0,0,0}$t$}%
}}}}
\put(3826,-1186){\makebox(0,0)[b]{\smash{{\SetFigFont{10}{12.0}{\familydefault}{\mddefault}{\updefault}{\color[rgb]{0,0,0}$t'$}%
}}}}
\put(4051,-736){\makebox(0,0)[lb]{\smash{{\SetFigFont{10}{12.0}{\familydefault}{\mddefault}{\updefault}{\color[rgb]{0,0,0}Background label}%
}}}}
\put(4051,-196){\makebox(0,0)[lb]{\smash{{\SetFigFont{10}{12.0}{\familydefault}{\mddefault}{\updefault}{\color[rgb]{0,0,0}Part-shape labels}%
}}}}
\put(3241,-1186){\makebox(0,0)[b]{\smash{{\SetFigFont{10}{12.0}{\familydefault}{\mddefault}{\updefault}{\color[rgb]{0,0,0}$e\in E_a$}%
}}}}
\end{picture}%

%% file: bilder/compos.pdf_t
\begin{picture}(0,0)%
\includegraphics{compos.pdf}%
\end{picture}%
\setlength{\unitlength}{4144sp}%
\begingroup\makeatletter\ifx\SetFigFont\undefined%
\gdef\SetFigFont#1#2#3#4#5{%
  \reset@font\fontsize{#1}{#2pt}%
  \fontfamily{#3}\fontseries{#4}\fontshape{#5}%
  \selectfont}%
\fi\endgroup%
\begin{picture}(5416,1456)(533,-924)
\put(3961,209){\makebox(0,0)[lb]{\smash{{\SetFigFont{10}{12.0}{\familydefault}{\mddefault}{\updefault}{\color[rgb]{0,0,0}$K^2$}%
}}}}
\put(2521,-691){\makebox(0,0)[rb]{\smash{{\SetFigFont{10}{12.0}{\familydefault}{\mddefault}{\updefault}{\color[rgb]{0,0,0}$K^1$}%
}}}}
\put(2071,-241){\makebox(0,0)[b]{\smash{{\SetFigFont{10}{12.0}{\familydefault}{\mddefault}{\updefault}{\color[rgb]{0,0,0}$\Rightarrow$}%
}}}}
\put(2521,-241){\makebox(0,0)[rb]{\smash{{\SetFigFont{10}{12.0}{\familydefault}{\mddefault}{\updefault}{\color[rgb]{0,0,0}$b^1$}%
}}}}
\put(4411,-241){\makebox(0,0)[b]{\smash{{\SetFigFont{10}{12.0}{\familydefault}{\mddefault}{\updefault}{\color[rgb]{0,0,0}$\Leftarrow$}%
}}}}
\put(3961,-241){\makebox(0,0)[lb]{\smash{{\SetFigFont{10}{12.0}{\familydefault}{\mddefault}{\updefault}{\color[rgb]{0,0,0}$b^2$}%
}}}}
\end{picture}%